\documentclass{article}

\usepackage{PRIMEarxiv}

\usepackage[T1]{fontenc}
\usepackage{graphicx}
\graphicspath{./}
\usepackage{amsmath}
\usepackage{amssymb}
\usepackage{amsfonts}
\usepackage{times}  
\usepackage[hyphens]{url}  
\usepackage{graphicx} 
\usepackage{multicol}
\usepackage{multirow}
\usepackage{booktabs}
\usepackage{soul}
\usepackage{tikz}
\usepackage{caption}
\usepackage{subcaption}
\usepackage{rotating}

\usepackage[ruled,vlined]{algorithm2e}
\usepackage[pagebackref,breaklinks,colorlinks]{hyperref}
\usepackage{color}

\DeclareMathOperator*{\argmax}{argmax}
\DeclareMathOperator*{\softmax}{softmax}
\DeclareMathOperator*{\expectation}{\mathbb{E}}

\def\checkmark{\tikz\fill[scale=0.4](0,.35) -- (.25,0) -- (1,.7) -- (.25,.15) -- cycle;} 

\title{Correlation Weighted Prototype-based Self-Supervised One-Shot Segmentation of Medical Images}
\author{Siladittya Manna\\
Indian Statistical Institute, Kolkata\\
\texttt{siladittya\_r@isical.ac.in}
\And
Saumik Bhattacharya\\
Indian Institute of Technology Kharagpur\\
\texttt{saumik@ece.iitkgp.ac.in}
\And
Umapada Pal\\
Indian Statistical Institute, Kolkata\\
\texttt{umapada@isical.ac.in}
}
\begin{document}
\maketitle              
\begin{abstract}

Medical image segmentation is one of the domains where sufficient annotated data is not available. This necessitates the application of low-data frameworks like few-shot learning.  Contemporary prototype-based frameworks often do not account for the variation in features within the support and query images, giving rise to a large variance in prototype alignment. In this work, we adopt a prototype-based self-supervised one-way one-shot learning framework using pseudo-labels generated from superpixels to learn the semantic segmentation task itself. We use a correlation-based probability score to generate a dynamic prototype for each query pixel from the bag of prototypes obtained from the support feature map. This weighting scheme helps to give a higher weightage to contextually related prototypes. We also propose a quadrant masking strategy in the downstream segmentation task by utilizing prior domain information to discard unwanted false positives. We present extensive experimentations and evaluations on abdominal CT and MR datasets to show that the proposed simple but potent framework performs at par with the state-of-the-art methods.

\end{abstract}
\section{Introduction}
Semantic Segmentation is one of the critical applications in computer vision. 
Applications of semantic segmentation to medical image analysis for assisting medical personnel in disease diagnosis are also plenty. For efficient and reliable analysis of medical images, contemporary deep-learning methods require large-scale datasets annotated by expert medical personnel. However, unlike natural image datasets, obtaining annotated high-quality medical image datasets is time-consuming and labour-intensive. To avoid the scarcity of large-scale datasets in medical image analysis, the few-shot learning paradigm has gained popularity among researchers.

In this work, we adopt the few-shot learning approach for learning to segment organs from MR or CT query scans, from a limited number of given support MR or CT slices and their corresponding ground truth segmentation masks. Depending on the pipeline, few-shot segmentation frameworks can be primarily of two types: prototype feature learning and affinity learning \cite{li2021asgnet}. Prototype feature learning consists of constructing prototypes utilizing the support image and the support mask information. Each prototype represents a defined spatial region in the support image. These prototypes are used to find pixels in the query image which are similar to them and are scored accordingly to segment it into foreground and background. Prototype-based features are more robust to noise than pixel-based features \cite{li2021asgnet}. Prototypical methods also drop spatial information, which is important when the variation between support and query images is considerably significant \cite{li2021asgnet}. Prototypical methods are also responsible for losing discriminability because of the masked pooling process to generate prototypes \cite{li2021asgnet}. To address this issue, we create prototypes for foreground and background pixels, which preserve the contextual spatial information required for effective discrimination between the foreground and background pixels. To prevent loss of information, we adopt a correlation-weighted aggregation approach such that the information of all the prototypes corresponding to foreground or background is present in the aggregated prototype.

In PANet \mbox{\cite{wang2019panet}}, a prototypical alignment-based strategy was proposed, wherein the masked support image embedding is mapped to the feature space, and the query mask is predicted by matching the query prototypes to the nearest prototype in the embedding space. However, PANet resorts to a global masked pooling operation, which is not suitable for medical image segmentation, as it can result in the loss of spatial orientation information. In ALPNet \mbox{\cite{ouyang2022alpnet}}, which is also a prototype-based framework, a local prototype-based approach is adopted to preserve local information using an adaptive local prototype pooling framework. However, such an approach ignores the global contextual information. In this work, we generate a single prototype for each pixel in the query feature map, that encodes both the global and local spatial context. We resort to a correlation-based aggregation approach, where the prototypes which are similar to a particular query pixel or located close in terms of spatial context, as well as its neighbourhood, will get a higher score than prototypes that are located far away in terms of spatial context. The probability scores are used to generate a weighted prototype for each query pixel. The correlation-based probability weighting scheme allows dynamic prototype generation for each query pixel by giving more weight to contextually related prototypes.
In addition to the framework design mentioned above, we also utilize prior domain information to further reduce the effect of false positives in the final downstream task by using \textit{quadrant masking} scheme.

The main contributions of our work are as follows.
\begin{itemize}
    \item We propose a novel correlation-weighted prototype aggregation-based self-supervised one-way one-shot learning framework for the segmentation of organs from abdominal magnetic resonance or computed tomography scans.
    \item We also propose a prior domain knowledge-informed quadrant masking scheme for discarding false positives in medical image segmentation tasks.
    \item Extensive experimental evidence on two datasets on abdominal magnetic resonance imaging and computed tomography shows the efficacy of the proposed simple but potent method.
\end{itemize}

\section{Related Work}

\subsection{Few Shot Segmentation}

One of the first pioneering works in Few Shot Segmentation (FSS) was presented in \cite{shaban2017oslss}, where a conditioning branch was used to predict weights, which serves as classifier weights for the query image feature obtained from the segmentation branch. The idea was extended in \cite{Rakelly2018ConditionalNF} using sparse positive and negative support pixels. In \cite{Rakelly2018FewShotSP}, a guided network is used to utilize information from the latent representation from the FSS task to segment query pixels. This work was further improved and extended in \cite{zhang2020guidednetcrf,MennatullahSiam2019wtimp,siam2019amp,bhunia2019logoret}.

\cite{Dong2018FewShotSS} presents one of the first works in the Prototypical Learning paradigm, by fusing prototypes from support images with the query image features using similarity scores. A leap in the paradigm of prototype learning was shown in \cite{wang2019panet}, where the prototype alignment strategy was introduced for maintaining cyclic consistency between the ground truth and the predicted segmentation mask inducing regularizing effect in training. Instead of altering the input structure as in \cite{Dong2018FewShotSS,Rakelly2018ConditionalNF}, the authors in \cite{zhang2020sgone} adopted a separate segmentation guidance framework based on similarity. \cite{liu2020papnfss} argue that the previous prototype-based methods do not take into account the various appearance of different parts in an object and propose a prototype-based part-aware framework to capture rich and fine-grained features. \cite{yang2020pmmfss} also pointed out that the primary disadvantage of existing prototype-based methods is the pooling operations which destroy the spatial layout information of the objects, and thereby proposed a prototype mixture model to solve the semantic ambiguity in prototype-based models.

To preserve the spatial correspondence between support and query image pixels, \cite{liu2022fssotmmf} uses a partial optimal transport-based matching. A multi-level variation of the same was done in \cite{yang2020dan}. \cite{li2021asgnet} and \cite{fan2022selfsupp} also aim to solve the same problem. \cite{zhang2021sgcgfss} and \cite{liu2022dpcn} aim to capture the intrinsic details to improve segmentation quality. \cite{haoyu2021poshp} and \cite{chen2022apanet} attempts to reduce testing bias in the FSS setting. An attempt to improve the discrimination between similar classes is presented in \cite{okazawa2022ipr}.

\subsection{Self-Supervised Segmentation}

The application of self-supervised learning frameworks in segmentation, although limited, follows two paths. One where the pre-training task is different from the downstream task of segmentation, and the other where both are the same. Works like \cite{guizilini2013ossdo,singh2018overheadimg}, \cite{ji2019iic,chen2019sslmiaicr,Zhu2020SelfSupervisedTF,ouali2020autouis,hoyer2021threeways,gao2022urltissseg} uses a pre-training stage to learn representations from the base dataset and then utilises the representation for a downstream semantic segmentation task. \cite{gansbeke2021maskcontrast} uses unsupervised saliency to generate object proposals and then optimizes a contrastive learning objective on the features obtained from the proposals to learn representations for semantic segmentation. For the second type, pseudo-masks are used as the segmentation masks in the pre-training stage. The above strategy is adopted in \cite{ouyang2020superpixels,araslanov2021ssaugcon,ouyang2022alpnet,amac2022masksplit}. In \cite{araslanov2021ssaugcon}, the authors use the output masks of a momentum update net as target pseudo-masks. In \cite{ouyang2020superpixels,ouyang2022alpnet}, the pseudo-masks are generated using the Felzenszwalb algorithm \cite{felzenszwalb2004imgseg} and the model using a few-shot learning strategy, the query image is an augmented version of the support image itself. 

In our work, to address the issue of false positives and also to preserve the spatial layout information \cite{yang2020pmmfss}, we propose a dynamic prototype-based framework that will weigh the background prototypes according to correlation with the pixel features.

\section{Methodology}

\begin{figure}[!ht]
    \centering
    \includegraphics[width = 0.95\linewidth]{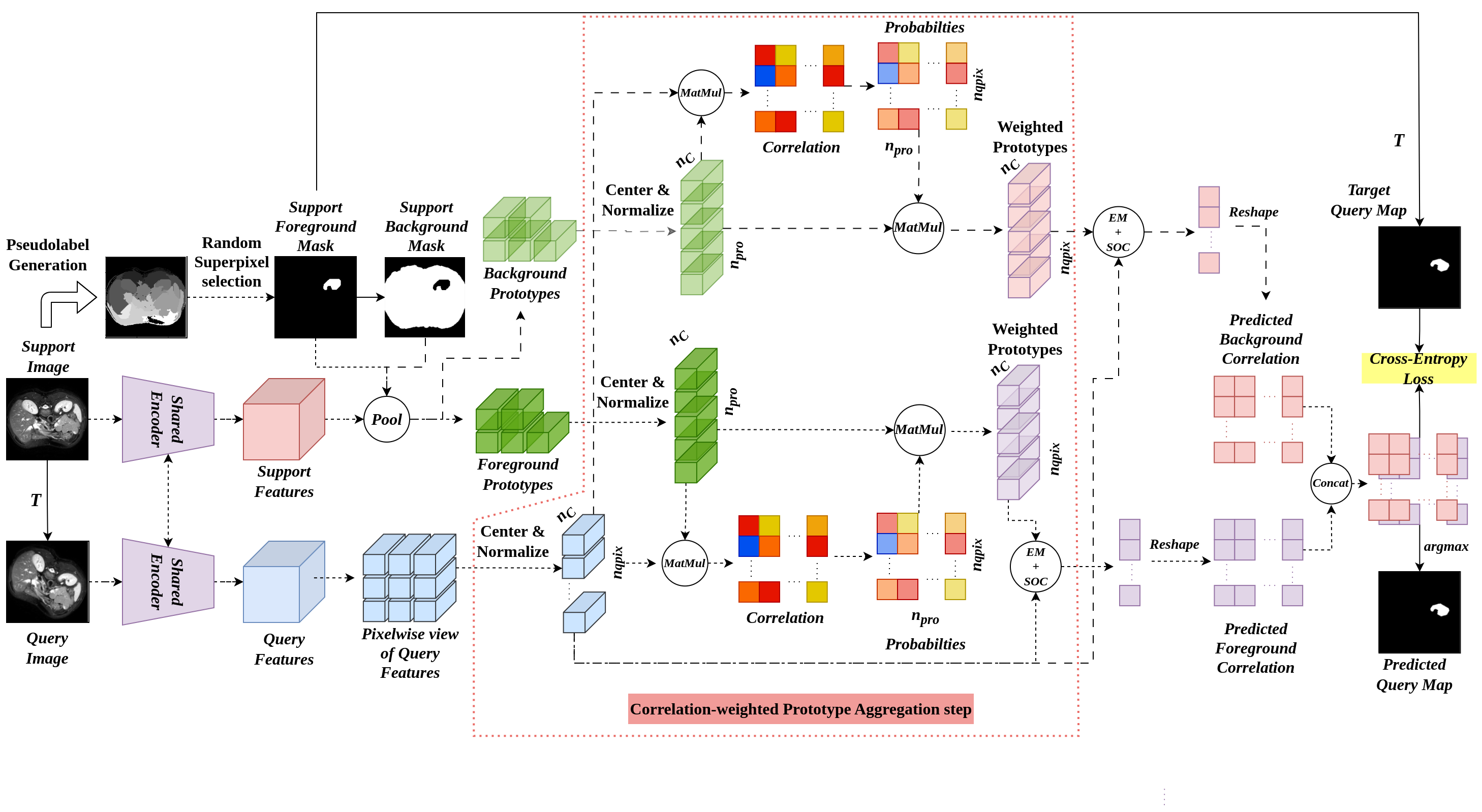}
    \caption{The figure depicts the entire working principle of the proposed framework. For clarity, we have also indicated the novel proposed correlation-weighted prototype aggregation step using a dotted red bounding box. \textit{\textbf{T}} indicates the transformation applied to the support image to generate the query image only in the pre-training stage. \textit{\textbf{Pool}} denotes pooling the feature map of the region denoted by the mask. \textit{\textbf{MatMul}} denotes Matrix Multiplication. `\textit{\textbf{EM+SOC}}' denotes Element-wise Multiplication and Sum over Channels. \textit{\textbf{Concat}} denotes the concatenation operation. (Best viewed at 300\%)}
    \label{fig:cowpro}
\end{figure}


\subsection{Problem Definition}

In the few-shot learning framework, the dataset is split into two parts, training dataset $\mathcal{D}_{train}$ and testing dataset $\mathcal{D}_{test}$. In both training and testing datasets, each sample consists of the input and the associated ground truth, $(\mathcal{X}, \mathcal{Y})$. In our work, $\mathcal{X}$ and $\mathcal{Y}$ correspond to slices from the abdominal MR or CT scans and the associated superpixel pseudo-masks generated in the pre-training stage, respectively. Whereas, in the evaluation or testing stage, the original ground truth masks for each organ are used with the MR or CT slices. Furthermore, no overlap should be present between the classes present in $\mathcal{D}_{train}$ and $\mathcal{D}_{test}$.

In the few-shot learning framework, we need to consider two sets of data, the \textit{Support} set $\mathcal{S}$ and the \textit{Query} set $\mathcal{Q}$. The \textit{Support} set $\mathcal{S}$ consists of the tuple $\{ \mathcal{X}^i_s, \mathcal{Y}^i_s(l) \}^k_{i=1}$, where $\mathcal{X}^i_s$ is the $i$-th sample in the Support set with the segmentation mask $\mathcal{Y}^i_s(l)$ for the class $l$, where $l$ belongs to the set of novel classes available during the testing phase. The primary objective is to learn an approximate function $f$ which takes as input the support set $\mathcal{S}$ and the query image $\mathcal{X}_q$ and predicts the binary mask $\hat{\mathcal{Y}_q}$ of the unseen classes in $\mathcal{X}_q$, denoted by the support mask $\mathcal{Y}_s(l)$. 

The support set $\mathcal{S}$ is a subset of $\mathcal{D}_{train}$. During training, the input to the model is $(\mathcal{S}, \mathcal{X}_q)$. Such a pair is called an \textit{episode}. If during training, the value of $k$ is $1$, that is, we use only a single image in the support set, then the learning is known as one-shot learning, which we adopt in this work. If $k > 1$, it is known as few shot learning. If the number of classes is $N$, then we call it $N$-way $k$-shot learning.

\subsection{Overview of the Proposed Approach}

We propose a self-supervised approach for one-shot learning of segmentation in MR and CT scans. The skeleton of our framework is based on a prototype-based segmentation strategy. For the first step, we use superpixel-based pseudo-segmentation mask generation. 
In our work, we adopt a dynamic prototype generation approach, one for each query feature map pixel. The dynamic nature of the prototype is the result of the aggregation step using correlation-based probability scores. The final score is obtained by calculating the correlation score of each query pixel to their assigned prototype.

In the downstream segmentation phase, we utilize the slide index information as in \cite{ouyang2020superpixels}, to filter out false positives (FP) obtained from other organs or regions on abdominal scans. Furthermore, we also propose to use the spatial location information of the respective organ $(l)$ to segment. 

\subsection{Generation of Pseudo Segmentation Masks}
\label{subsec:genpseudomask}

We follow the same strategy as in \cite{ouyang2020superpixels} for the generation of pseudo-segmentation masks. As stated in \cite{ouyang2020superpixels}, superpixel-based segmentation satisfies two properties: for each class, the representations should be clustered to be discriminative under a similarity metric, and the representations should also be invariant across images to ensure robustness. Otherwise, the regions that denote the same class in the support and query images would not be mapped together in the feature space. A superpixel-based clustering strategy ensures that regions with similar pixel features are clustered together. This ensures consistency over each of the pseudo-labels as well. 

For every slice in abdominal scans (say $\mathcal{X}_i$), the Felzenszwalb image segmentation algorithm \cite{felzenszwalb2004imgseg} $\mathcal{F}$ is applied to the slice to generate the super pixels, $\mathcal{S}_{p} = \mathcal{F}[\mathcal{X}^i]$). During self-supervised training, a superpixel is randomly chosen, $l_s \sim \mathcal{U}[0,|\mathcal{S}_p|-1]$ and converted to a binary mask to be used as the segmentation mask. A sample of the superpixels obtained is shown in Fig. \ref{fig:cowpro}.

\begin{equation}
    \mathcal{Y}_s = [\mathcal{S}_p \in \{l_s\}]
\end{equation}


\subsection{Feature Extraction}

Each episode $(\mathcal{S}, \mathcal{X}_q)$ is passed through the encoder $f_{\theta}$, which gives us the support and query feature maps, which we denote by $f_{\theta}(\mathcal{X}_s)$ and $f_{\theta}(\mathcal{X}_q)$. In our case, the encoder takes an input of dimension $3 \times 256 \times 256$ and outputs a feature map with dimensions $256 \times 32 \times 32$. We use the $deeplab\_v3$ version of ResNet101 available from the $torchvision$ library. To ensure that the output dimensions match the specifications mentioned above, we used $dilation$ in the last two layers of the encoder, similar to \cite{ouyang2022alpnet}.

\subsection{Correlation Weighted Prototype Aggregation}
\label{subsec:corretprotoagg}
The principle component of our proposed framework is the prototype aggregation module. This novel correlation-weighted prototype-aggregation module primarily consists of four steps: 1) Prototypes Extraction, 2) Correlation computation, 3) Probability score computation, and 4) Prototype Aggregation. 
The prototype aggregation steps are done separately for foreground and background. 

\subsubsection{Prototype Extraction}
\label{subsubsec:protoext}
We do not extract the foreground features by merely doing global average pooling using the support mask $\mathcal{Y}_s$. In this case, we follow the steps described in \cite{ouyang2020superpixels}, for extracting the foreground and background prototypes. The first step to obtaining the foreground (or background) prototypes is to downsample the segmentation mask to spatial dimension $H \times W$ using an average pooling operation with a window of spatial dimensions $4 \times 4$. However, using an average pooling operation may result in values that are not binary (0 or 1). To get a downsampled binary mask, we threshold the interpolated mask. For the foreground, we select a threshold that is $0.8$ times the maximum value of the downsampled mask. For the background, we use a threshold that is equal to the mean of the downsampled mask, following \cite{wu2022dclass}. 
\begin{equation}
    \mathcal{Y}_{s(H \times W)} = [AvgPool_{4 \times 4}(\mathcal{Y}_s) > \tau]
\end{equation}

where $AvgPool$ refers to the Average Pooling operation applied on the binary mask $\mathcal{Y}_s$, and $\tau$ is the threshold. However, we find that, for the label sets containing Liver and Spleen, using a threshold of 0.95 for both foreground and background worked better than the aforesaid thresholds (See Table \ref{tab:threshtab}). Next, the locations $w$ where the downsampled binarized mask $\mathcal{Y}_{s(H \times W)}$ is non-zero are processed. The pixels in the support feature map $f_{\theta}(\mathcal{X}_s) \in \mathbb{R}^{D \times H \times W}$, whose locations match those in $w$, are chosen as prototypes. For the foreground prototypes, we also include the global prototype with the obtained prototypes to avoid an empty set of prototypes resulting from the averaging over the small area of the foreground, following \cite{ouyang2022alpnet}.
\begin{equation}
    \mathcal{P} = f_{\theta}(\mathcal{X}_s)[\mathcal{Y}_{s(H \times W)} \in \{1\}]
\end{equation}

Before calculating the cosine similarity between the prototypes $\mathcal{P} \in \mathbb{R}^{D \times N_{pro}}$ and the pixels of the query feature map $f_{\theta}(\mathcal{X}_q)$, we subtract the mean of each of the $N_{pro}$ prototypes along the channel dimension. 
\begin{equation}
    \mathcal{P}^j = \mathcal{P}^j - \frac{1}{D}\sum_{d=1}^D \mathcal{P}^j[d]
\end{equation}

where $\mathcal{P}^j$ is the $j$-th prototype. The steps mentioned above are done for both the foreground and background prototypes. To obtain the foreground prototypes, we simply take $\mathcal{Y}_s$ as the foreground mask $\mathcal{Y}_s^{FG}$, whereas for the background prototypes, we take $\mathcal{Y}_s^{BG} = 1 - \mathcal{Y}^{FG}_s$ as the background mask.

\subsubsection{Query Features Centering}

The same mean subtraction operation is also done for the query pixels in the output feature map, as follows,

\begin{equation}
    f_{\theta}(\mathcal{X}_q)_{h,w} = f_{\theta}(\mathcal{X}_q)_{h,w} - \frac{1}{D}\sum_{d=1}^D f_{\theta}(\mathcal{X}_q)_{h,w}[d]
\end{equation}

where, $\{h,w\}$ denotes the location of the pixels in the feature map. 

\subsubsection{Correlation computation}
Having obtained the query feature map $f_{\theta}(\mathcal{X}_q)$ of dimensions $D \times H \times W$ and prototypes $\mathcal{P}$ of dimensions $D \times N_{pro}$, we proceed to compute the cosine similarity between these entities. This results in a correlation matrix or a cosine similarity matrix $\mathcal{C}$, which has dimensions $N_{pro} \times H \times W$. This 3D matrix represents the correlation score of all the prototypes obtained in the previous step for each pixel in the query feature map. 
\begin{equation}
    \mathcal{C}(j,h,w) = f_{\theta}(\mathcal{X}_q)(h,w) \odot \mathcal{P}^j
\end{equation}

where $\odot$ indicates the operation of the dot product, $f_{\theta}(\mathcal{X}_q)(h,w)$ denotes the feature at the location $(h,w)$, and $\mathcal{P}^j$ is the $j$-th prototype. $\mathcal{C}(j,h,w)$ has dimensions $N_{pro} \times H \times W$. The correlation score indicates how similar each prototype is to the query feature map pixels. Intuitively, a prototype $p \in \mathcal{P}$ with a higher correlation score with a pixel on the query feature map $f_{\theta}(\mathcal{X}_q)(h,w)$ can be said to be more similar than a prototype with a lower correlation score, in terms of feature similarity. As the feature extractor uses dilation, the receptive field of each pixel in the output feature map has a very large receptive field. Hence, contextual or neighbourhood information is encoded in each foreground prototype $\mathcal{P}^{FG}$. This contextual information will help distinguish the region indicated by the support mask $\mathcal{Y}_s$ and the other regions. 

\subsubsection{Probability score computation}

The probability of each prototype being similar to a particular query pixel is calculated by taking softmax over the prototypes as follows:
\begin{equation}
    \mathcal{M}_{prob}(h,w) = \text{softmax}_{j \in \mathcal{P}} [\mathcal{C}(h,w) / t]
\end{equation}

where $\mathcal{M}_{prob}(h,w)$ denotes the probability of the prototypes $\mathcal{P}$ with respect to the query pixel at the location $(h,w)$, and $t$ is a temperature parameter. $\mathcal{M}_{prob}(h,w)$ has dimensions $N_{pro} \times H \times W$.

When calculating the scores for the background prototypes $\mathcal{P}^{BG}$, the background query pixels which are spatially close to a particular background query pixel (say, $f_{\theta}(\mathcal{X}_q)(h,w)$), will yield different correlation scores. This may result in erroneous predictions or increased false positives if we weigh all the prototypes equally. The background prototypes which are spatially farther from $f_{\theta}(\mathcal{X}_q)(h,w)$ region or feature-wise dissimilar will bring the final score down, thereby increasing false positives. This requires a dynamic prototype that captures contextual information effectively. This is made possible by giving a probabilistic weightage to contextually similar prototypes.

\begin{figure}[!ht]
     \centering
     \begin{subfigure}[b]{0.45\linewidth}
         \centering
         \includegraphics[width=0.9\linewidth]{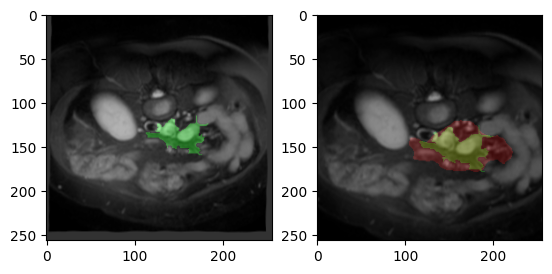}
         \caption{Iteration: 25000}
         \label{fig:tr1}
     \end{subfigure}
     \hfill
     \begin{subfigure}[b]{0.45\linewidth}
         \centering
         \includegraphics[width=0.9\linewidth]{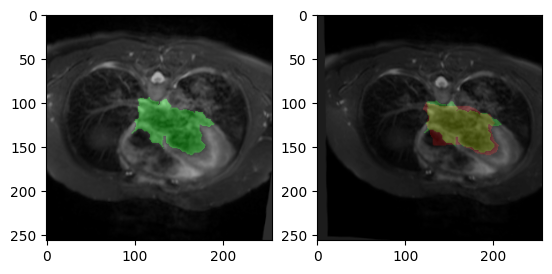}
         \caption{Iteration: 50000}
         \label{fig:tr2}
     \end{subfigure}
     \\
     \begin{subfigure}[b]{0.45\linewidth}
         \centering
         \includegraphics[width=0.9\linewidth]{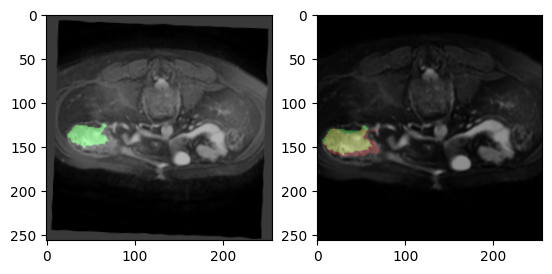}
         \caption{Iteration: 75000}
         \label{fig:tr3}
     \end{subfigure}
     \hfill
     \begin{subfigure}[b]{0.45\linewidth}
         \centering
         \includegraphics[width=0.9\linewidth]{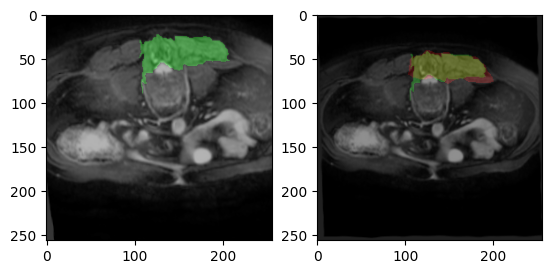}
         \caption{Iteration: 100000}
         \label{fig:tr4}
     \end{subfigure}
        \caption{Predictions in training phase at 25K, 50K, 75K, 100K iterations. The left image in Figs. \ref{fig:tr1}-\ref{fig:tr4} is the support image $\mathcal{X}_s$ and the support mask is denoted in \textit{green}. The right image in Figs. \ref{fig:tr1}-\ref{fig:tr4} is the query image. The ground truth is denoted by \textit{green} and the predicted mask is indicated by \textit{ red}. (Use 300\% zoom for better visibility)}
        \label{fig:trainingimgs}
\end{figure}

\subsubsection{Prototype Aggregation}

The weighting scheme necessary for a dynamic and contextual prototype generation is done by aggregating the prototypes on the basis of probability scores obtained from the correlation values between the prototypes and the query pixels. The aggregated dynamic prototype is obtained by a weighted average of all the prototypes using the probabilities obtained in the previous step, as follows:

\begin{equation}
    \mathcal{P}_{agg}(h,w) = \sum_{j = 1}^{N_{pro}} \mathcal{M}_{prob}(h,w) \cdot \mathcal{P}^j
\end{equation}

where $\mathcal{P}_{agg}(h,w) \in \mathbb{R}^{D \times 1 \times 1}$ denotes the aggregated prototype for the query pixel at location $(h,w)$, $\mathcal{P}^j \in \mathbb{R}^{D \times 1 \times 1}$ denotes the $j$-th prototype.

\subsection{Mask Prediction}
\subsubsection{Computing the final score}

Once we have the aggregated prototype, we can compute the final scores for each query pixel. The final score is calculated by simply calculating the cosine similarity of the aggregated prototype $\mathcal{P}_{agg}(h,w)$ with the pixel feature of the query in location $(h,w)$, as follows.

\begin{equation}
    s_{FG}(h,w) = \mathcal{P}_{agg}^{FG}(h,w) \odot f_{\theta}(\mathcal{X}_q)(h,w)
\end{equation}

\begin{equation}
    s_{BG}(h,w) = \mathcal{P}_{agg}^{BG}(h,w) \odot f_{\theta}(\mathcal{X}_q)(h,w)
\end{equation}

where $s_{FG}(h,w)$ and $s_{BG}(h,w)$ are the scores for the query pixels with respect to the foreground and background prototypes, respectively, and $\mathcal{P}_{agg}^{FG}$ and $\mathcal{P}_{agg}^{BG}$ are the aggregated prototypes for the foreground and background, respectively.

\subsubsection{Final Prediction}

The final prediction is obtained by choosing the class with the highest probability or the similarity scores for the foreground and background for each query pixel. Thus, the final prediction for each query pixel is obtained as follows:

\begin{equation}
    \hat{\mathcal{Y}_q}(h,w) = \argmax_{\{BG,FG\}} \softmax_{\{BG,FG\}}[s_{BG}, s_{FG}]
\end{equation}

where $\hat{\mathcal{Y}_q}(h,w)$ is the predicted query mask. A few examples of the query predictions and the associated ground truth from the training stage are shown in Fig. \ref{fig:trainingimgs}.

\subsection{Training Pipeline}

In each iteration $t$, we take an episode $((\mathcal{X}_s,\mathcal{Y}_s), \mathcal{X}_q)$ as input, and the model predicts $\hat{\mathcal{Y}_q}$ as output. The query image $\mathcal{X}_q$ is obtained by applying geometric and intensity transformations on the support image $\mathcal{X}_s$, that is, $\mathcal{X}_q = \mathcal{T}_{geo}(\mathcal{T}_{int}(\mathcal{X}_s))$. The pseudo-ground truth is obtained by only applying the geometric transformation $\mathcal{T}_{geo}$ on the support mask $\mathcal{Y}_s$ for a randomly chosen pseudo-superpixel class. 

Geometric transformations $\mathcal{T}_{geo}$ consist of affine and elastic transformations to emulate the changing shapes of the class labels in the downstream task. Intensity transformations $\mathcal{T}_{int}$ consist of gamma transformation to account for the varying intensity of the pixels between scans of different patients. The parameters for the geometric and intensity transformation are the same as used in \cite{ouyang2020superpixels}.

Since the encoder $f_{\theta}$ output is of spatial dimension $32 \times 32$, we interpolate the final prediction to $256 \times 256$ before calculating the loss using bilinear interpolation. To optimize the model parameters, we minimize the cross-entropy loss $\mathcal{L}^t_{ssl}$ for all the query pixels.
\begin{equation}
    \mathcal{L}^t_{ssl}(\theta) =  - \expectation_{h,w}\left[ \lambda_{h,w} \mathcal{T}_{geo}(\mathcal{Y}^t_s)(h,w) \log \left( \hat{\mathcal{Y}^t_q}(h,w) \right) \right]
\end{equation}
where $\hat{\mathcal{Y}^t_q}(h,w)$ is the predicted output of $\mathcal{T}_{geo}(\mathcal{Y}^t_s)(h,w)$ taking $\mathcal{X}_q = \mathcal{T}_{geo}(\mathcal{T}_{int}(\mathcal{X}_s))$ as query. $(h,w)$ denotes the location in the predicted query mask or pseudo-ground-truth mask. $\lambda_{h,w}$ denotes the class weight applied during training.

Similar to \cite{ouyang2020superpixels}, we also adopt the Cyclic Consistency Regularization (CCR) following \cite{wang2019panet}. To implement CCR, we interchange query and support images. For the support mask, we use the predicted query output $\hat{\mathcal{Y}_q}$ as the foreground mask, and the support mask $\mathcal{Y}_s$ in the forward iteration is used as the pseudo-ground-truth. The episode in the CCR step consists of $((\mathcal{X}_q, \hat{\mathcal{Y}_q}),\mathcal{X}_s)$. In the CCR step, we use an initial threshold of 0.95 in the prototype extraction step to filter out noisy predictions, following which the aforementioned steps are followed. Otherwise, we see a drop in performance by about 2\% in dice score. The CCR loss is represented as
\begin{equation}
     \mathcal{L}^t_{reg}(\theta) =  - \expectation_{h,w}\left[\mathcal{Y}^t_s(h,w) \log \left( \hat{\mathcal{Y}^t_s}(h,w) \right) \right]
\end{equation}

where $\hat{\mathcal{Y}^t_s}(h,w)$ is the predicted output of $\hat{\mathcal{Y}^t_q)(h,w)}$ taking $\mathcal{X}_s$ as a query.

Hence, the total loss is as follows,
\begin{equation}
    \mathcal{L}^t = \mathcal{L}^t_{ssl} + \mathcal{L}^t_{reg} 
\end{equation}

To handle the imbalance, we set the class weights at $0.05$ for the background pixels or the class label $0$, and a weight of $1.0$ for the foreground pixels or the class label $1$.

Furthermore, it is to be noted that during training, we divide the class labels in abdominal CT or MR 
into two parts, namely, upper abdomen consisting of \textit{right kidney} and \textit{left kidney}, and lower abdomen consisting of \textit{liver} and \textit{spleen}. When training on slices from the upper abdomen, we do not include slices containing lower abdomen classes and vice versa.

\subsection{Validation without Fine-tuning}

Following \cite{ouyang2020superpixels,ouyang2022alpnet}, we evaluate our model on a validation split, without further fine-tuning on the whole dataset. Although we don't fine-tune the model, we do use the class label information from the dataset. For a class label $l$, we only take the slices $[z_{min}, z_{max}]$ in which $l$ is present for the final predictions. 

\subsubsection{One-Shot Segmentation}

The evaluation strategy follows a one-shot segmentation task. Among the scans in the validation split, the last scan (if arranged in order) is selected as the support scan. The range of slices $[z_{min}, z_{max}]$ in both support and query scans is divided into 3 parts following the evaluation strategy followed in \cite{guharoy2020senet}. From the three support scan splits, the middle slice is selected as the support image for the whole of the corresponding query part. The evaluation step then follows the flow of the one-shot segmentation task, that is, a tuple $((\mathcal{X}^p_s, \mathcal{Y}^p_s), \mathcal{X}^{p,i}_q)$ is fed to the pre-trained model $f_{\theta}$ as input to obtain the predicted query mask $\hat{\mathcal{Y}^{p,i}_q}$, where $p$ is the part in which the slices belong and $i \in [z^q_{min},z^q_{max}]$ is the index of the slices of query scan in which the class label $l$ is present.

\subsubsection{Quadrant Masking Scheme}

During training, the classes in abdominal MRI or CT datasets are split into two parts, the upper abdomen (right and left kidneys) and the lower abdomen (liver and spleen). During validation, we divide each slice into quadrants. For each class label $l$, we identify which quadrants are occupied by it. The final predictions obtained from the one-shot segmentation step are masked such that the predictions from the quadrants in which the class label $l$ is present, are considered for the final metric calculation. For example, the class \textit{right kidney} is present in the left half of an MRI or CT slice. Hence, we mask the right half of each slice while making the final prediction. 
The quadrant masking scheme uses the quadrant information as a piece of soft prior information about the location of the target organ. To the best of our knowledge, no work has employed this scheme before. To fully understand the role of this quadrant masking scheme, we conduct an ablation study in Sec. \mbox{\ref{sec:ablquad}}, where we observe the significant effects of the soft prior knowledge in boosting the segmentation performance.

\subsubsection{Validation Metric}

To measure the performance of the proposed model, we use the Dice score as a metric, as is usually done in the medical image segmentation literature \cite{ouyang2020superpixels,ouyang2022alpnet,wang2019panet}.

\section{Experiments and Results}

\subsection{Implementation Details}

Training and evaluation were implemented using PyTorch. The training was done on a 24GB NVIDIA A5000 GPU. The average training time for each training run consisting of 100K iterations was about 4.5 hours. We used a batch size of 1. The initial learning rate of the SGD optimizer was set to $1e-3$ and decayed at $0.95$ per 1K iterations. 

\subsection{Datasets}

To demonstrate the effectiveness of the proposed approach, we diversified the input modalities by including both Magnetic Resonance (MR) and Computed Tomography (CT) scans in our work. 

For the Magnetic Resonance (MR) modality, we used the Combined Healthy Abdominal Organ Segmentation (CHAOS) Challenge (Task 5) from ISBI 2019 \cite{kavur2021chaos}. This dataset contains 20 3D T2-SPIR MRI scans. 

For the Computed Tomography (CT) modality, we used data from the 2015 MICCAI Multi-Atlas Abdomen Labeling Challenge (SABS). It contains abdominal scans from 30 patients.

For the experiments, we used a five-fold cross-validation setting, that is, in an experimental run, one-fifth of the dataset (a fold) is used as a validation set while the rest is treated as a training set.

\subsection{Results and Comparison}

\subsubsection{Quantitative Performance Analysis}
In this section, we present the results obtained by the proposed model on the two datasets: CHAOS and SABS. 
From Tables \ref{tab:chaosres} and \ref{tab:sabsres}, we can see that the proposed framework outperforms ALPNet \cite{ouyang2020superpixels} and also outperforms several current state-of-the-art methods in several classes. The bold font and the underlined text indicate the best and the second-best performance, respectively. The proposed algorithm outperforms CRAPNet on the CHAOS dataset and produces competitive results on the SABS dataset without any further fine-tuning of hyperparameters. This shows that the dynamic prototype aggregation technique improves the representation learning and generalizability of the model.

\begin{table}[!ht]
    \centering
    \caption{DICE score on Abdominal MR (CHAOS) Dataset. Reported Values are with Single Support Scan.}
    \begin{tabular}{|c|c|c|c||c|c||c|}
    \hline
        Method & Supervised & RK & LK & Liver & Spleen & Mean\\ \hline \hline
        SE-Net \cite{guharoy2020senet} & \checkmark &61.32 & 62.11 & 27.43& 51.80&50.66 \\ \hline
        Vanilla PANet \cite{wang2019panet} & \checkmark & 38.64& 53.45& 42.26& 50.90 &46.33 \\\hline
        ALPNet \cite{ouyang2020superpixels} & \checkmark & 58.99& 53.21& 37.32& 52.18& 50.43\\ \hline \hline
        SSL-PANet \cite{ouyang2020superpixels} & $\times$& 47.95 & 47.71 & 64.99 & 58.73 & 54.85\\ \hline
         SSL-ALPNet  \cite{ouyang2020superpixels}& $\times$& 78.39 & 73.63 & 73.05 & 67.02 & 73.03\\ \hline
         CRAPNet \cite{ding2023crapnet}& $\times$&  \textbf{82.77} & \underline{74.66} & \underline{73.82} & \underline{70.82} & \underline{75.52}\\ \hline \hline
         CoWPro (Ours) & $\times$& \underline{80.45} & \textbf{75.30} &  \textbf{75.77} & \textbf{71.51} & \textbf{75.56}\\ \hline \hline
    \end{tabular}
    \label{tab:chaosres}
\end{table}
\begin{table*}[!ht]
    \centering
     \caption{DICE score on Abdominal CT (SABS) Dataset.  Reported Values are with Single Support Scan.}
    \begin{tabular}{|c|c|c|c||c|c||c|}
    \hline
        Method & Supervised&RK & LK & Liver & Spleen & Mean\\ \hline \hline
        SE-Net \cite{guharoy2020senet} & \checkmark &14.34& 32.83& 0.27 &0.23& 11.91 \\ \hline
        Vanilla PANet \cite{wang2019panet}& \checkmark & 17.37& 32.34& 38.42& 29.59& 29.43\\ \hline
        ALPNet \cite{ouyang2022alpnet}& \checkmark  & 30.40& 34.96& 47.37& 27.73& 35.11 \\ \hline \hline
        SSL-PANet \cite{ouyang2020superpixels}& $\times$  & 34.69 & 37.58 & 61.71 & 43.73 & 44.42\\ \hline
         SSL-ALPNet  \cite{ouyang2020superpixels}& $\times$ & 54.82 & \underline{63.34} & \textbf{73.65} & 60.25 & 63.02\\ \hline
         CRAPNet \cite{ding2023crapnet}& $\times$  &  \textbf{67.33} & \textbf{70.91} & 70.45 & \textbf{70.17} & \textbf{69.72} \\ \hline \hline
         CoWPro (Ours)& $\times$   & \underline{58.99} & 62.66 & \underline{73.11} & \underline{67.97} & \underline{65.83}\\ \hline \hline
    \end{tabular}
    \label{tab:sabsres}
\end{table*}

\subsubsection{Qualitative Performance Analysis}

\begin{figure*}[!htb]
     \centering
     \begin{subfigure}[b]{0.245\linewidth}
         \centering
         \includegraphics[width=0.9\linewidth]{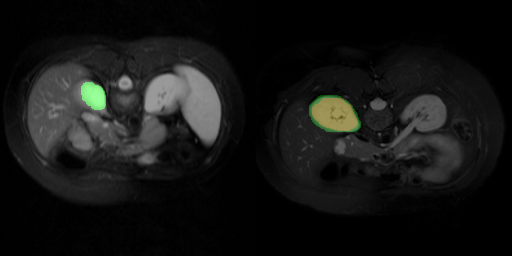}
         \caption{(MR) Right Kidney}
         \label{fig:tr1_}
     \end{subfigure}
     \hfill
     \begin{subfigure}[b]{0.245\linewidth}
         \centering
         \includegraphics[width=0.9\linewidth]{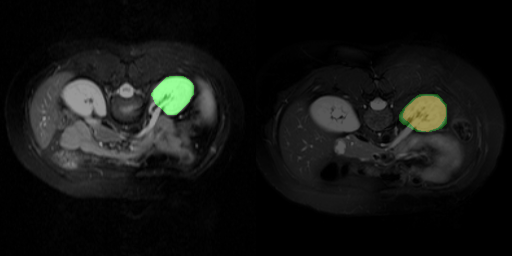}
         \caption{(MR) Left Kidney}
         \label{fig:tr2_}
     \end{subfigure}
     \hfill
     \begin{subfigure}[b]{0.245\linewidth}
         \centering
         \includegraphics[width=0.9\linewidth]{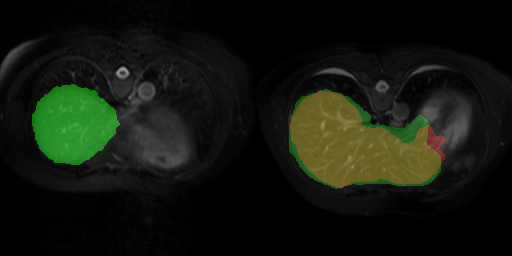}
         \caption{(MR) Liver}
         \label{fig:tr3_}
     \end{subfigure}
     \hfill
     \begin{subfigure}[b]{0.245\linewidth}
         \centering
         \includegraphics[width=0.9\linewidth]{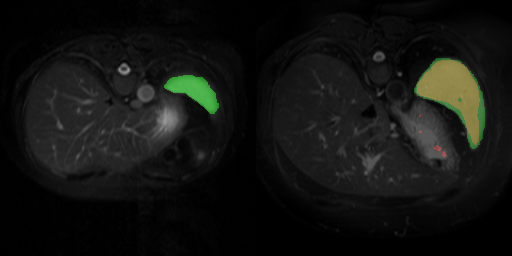}
         \caption{(MR) Spleen}
         \label{fig:tr4_}
     \end{subfigure}
     \\
     \begin{subfigure}[b]{0.245\linewidth}
         \centering
         \includegraphics[width=0.9\linewidth]{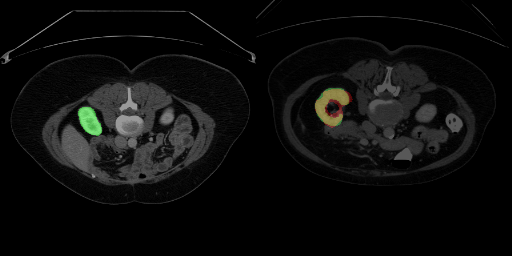}
         \caption{(CT) Right Kidney}
         \label{fig:tr5_}
     \end{subfigure}
    \hfill     
     \begin{subfigure}[b]{0.245\linewidth}
         \centering
         \includegraphics[width=0.9\linewidth]{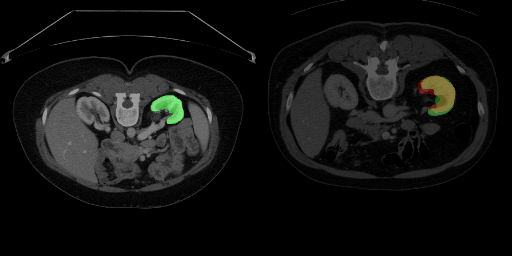}
         \caption{(CT) Left Kidney}
         \label{fig:tr6_}
     \end{subfigure}
     \hfill
     \begin{subfigure}[b]{0.245\linewidth}
         \centering
         \includegraphics[width=0.9\linewidth]{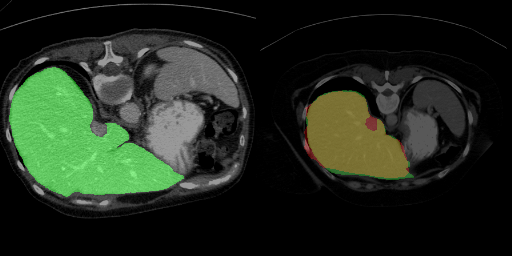}
         \caption{(CT) Liver}
         \label{fig:tr7_}
     \end{subfigure}
     \hfill
     \begin{subfigure}[b]{0.245\linewidth}
         \centering
         \includegraphics[width=0.9\linewidth]{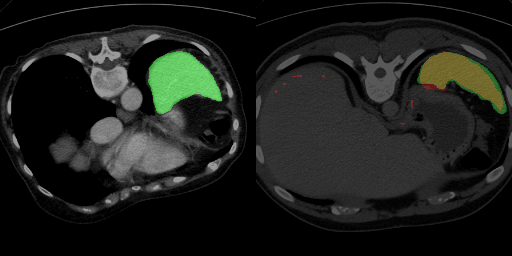}
         \caption{(CT) Spleen}
         \label{fig:tr8_}
     \end{subfigure}
        \caption{Figure showing the predictions obtained for 4 organs, Right Kidney, Left Kidney, Liver, and Spleen for two different modalities MR (CHAOS dataset) and CT (SABS dataset). (green) Ground Truth, (red) Prediction, (yellow) Ground Truth and Prediction overlap. (Use 300\% zoom for better visibility)}
        \label{fig:predictions}
\end{figure*}

The qualitative performance of the proposed model can be seen from the predictions presented in Fig. \ref{fig:predictions}. We can see the predictions for different organs on two different modalities compared to the ground truth. We can see that the model produces segmentation results close to the ground truth. 

\subsection{Ablation Studies}

\subsubsection{Studying the effect of Fixed vs. Dynamic Thresholding}

The effect of threshold on the performance of the proposed framework can be seen in Table \ref{tab:threshtab}, where we see the use of two different types of threshold, dynamic and fixed, for different sets of labels. In the CHAOS dataset, the use of a dynamic thresholding scheme similar to the one used in \cite{wu2022dclass} works better for the labels \textit{right kidney} and \textit{left kidney}, whereas a fixed threshold of 0.95, similar to \cite{ouyang2022alpnet} works better for the other two labels. However, for experiments on SABS, we do not see much variation in performance with the change in thresholds. The dynamic thresholding scheme assigns a threshold of 0.8 times the max value of the downsampled foreground mask, whereas for the background it uses the mean value of the downsampled background mask. The dynamic thresholding scheme allows the model to adapt to the local pixel intensities and infuses a local spatial information in the process. However, for large organs like the spleen and liver, the effect of dynamic thresholding is not positive.

\begin{table}[!htb]
    \centering
    \caption{Dice score obtained on abdominal CT and MR datasets using different thresholding schemes without Quadrant masking scheme.}
    \begin{tabular}{|c||c|c|c|c||c|c|c|c|}
    \hline
        \multirow{2}{*}{Threshold} &  \multicolumn{4}{c||}{Abdominal MR} & \multicolumn{4}{c}{Abdominal CT}\\ \cline{2-9}
        & R. Kidney & L. Kidney & Liver & Spleen & R. Kidney & L. Kidney & Liver & Spleen \\ \hline \hline
        Fixed & 79.06 & 72.24 & \textbf{75.04} & \textbf{71.19} & \textbf{58.99} & 62.47  & 73.19 & 66.67 \\ \hline
        Dynamic & \textbf{79.54} & \textbf{74.59} & 73.87 & 66.33 & 58.24 & \textbf{62.66}  & \textbf{73.71}& \textbf{67.97} \\ \hline
    \end{tabular}
    \label{tab:threshtab}
\end{table}

\begin{table}[!htb]
    \centering
    \caption{Dice score obtained on abdominal CT and MR datasets with and without quadrant masking scheme with fixed thresholding.}
    \begin{tabular}{|c||c|c|c|c||c|c|c|c|}
    \hline
        \multirow{2}{*}{Quadrant Masking} &  \multicolumn{4}{c||}{Abdominal MR} & \multicolumn{4}{c}{Abdominal CT}\\ \cline{2-9}
        & R. Kidney & L. Kidney & Liver & Spleen & R. Kidney & L. Kidney & Liver & Spleen \\ \hline \hline
        Yes & \textbf{79.66} & \textbf{75.30}  & \textbf{75.77} & \textbf{71.51}  & \textbf{58.99} & \textbf{63.31} & 73.18 & \textbf{66.85} \\ \hline
        No & 79.06 & 72.24 & 75.04 & 71.19 & \textbf{58.99} & 62.47 & \textbf{73.19} &  66.67 \\ \hline
    \end{tabular}
    \label{tab:ablquadtab}
\end{table}

\subsubsection{Effect of Quadrant Masking}
\label{sec:ablquad}

In Tab. \mbox{\ref{tab:ablquadtab}}, we observe the effect of the quadrant masking scheme on the segmentation performance on both the MR and CT datasets. Barring a few exceptions, the quadrant masking scheme has improved the dice score for all the organs. The primary reason behind the slight drop in performance can be attributed to the hard quadrant boundary assigned to the corresponding organs.

\subsubsection{Effect of number of aggregated prototypes}

The correlation-weighted prototype-aggregation step aims to incorporate the information from all the prototypes to prevent loss of information, as is generally the case in prototype-based methods. However, one may argue, that using all the prototypes may induce an unintended negative effect from anatomically different but semantically similar regions, consequently causing a degradation in performance. In this ablation study, we take the Top-100\%, 50\%, 10\%, 5\%, and 2\% most similar prototypes to predict foreground or background regions. This study also establishes the efficiency of our model in encoding contextual information, which is evident from the insignificant variation in the dice scores with the decrease in the number of prototypes. In Tab. \mbox{\ref{tab:abl2tab}}, we show the dice scores for varying numbers of prototypes in the inference stage for the left and right kidney over all the folds.

\begin{table}[!ht]
    \centering
    \caption{Dice score for $2 \times 2$ and $4 \times 4$ averaging window without Quadrant Masking.}
    \begin{tabular}{|cc|c|c|c|c|c|}
    \toprule
      \multicolumn{3}{|c|}{Averaging Window} & \multicolumn{2}{c|}{$2\times2$}  & \multicolumn{2}{c|}{$4\times4$} \\ \hline
      \multicolumn{3}{|c|}{Organ} & RK & LK & RK & LK \\ \hline \hline
     \parbox[t]{3mm}{\multirow{5}{*}{\rotatebox[origin=c]{90}{Percentage of}}} & \parbox[t]{3mm}{\multirow{5}{*}{\rotatebox[origin=c]{90}{prototypes}}}& 100\% & 79.99 & 76.22 & 80.77 & 72.87\\ 
        && 50\% & 79.89& 76.24& 80.72& 72.94\\
        && 10\% & 79.49& 75.22 & 80.94 & 73.81 \\
        && 5\% & 78.16& 73.46& 81.08 & 74.43 \\
        && 2\% & 75.69& 70.42& 80.68& 74.86\\
        \bottomrule
    \end{tabular}
    \label{tab:abl2tab}
\end{table}

\subsubsection{Effect of Averaging window}

In Tab. \mbox{\ref{tab:abl2tab}}, we observe the effect of changing the averaging window in the prototype generation step. We chose the left and right kidneys from the abdominal MR dataset to study the effect of the averaging window, as the two kidneys are almost similar in shape and size but vary only in the spatial context. We observe that using the same averaging window as in training, that is, an averaging window of $4 \times 4$, the performance of the proposed framework is better for the right kidney, than using an averaging window of $2 \times 2$. On the contrary, using an averaging window of $2 \times 2$ yields better performance than using an averaging window of $4\times 4$ on the left kidney. We believe that this discrepancy in the trend is primarily due to the different spatial contexts of the two organs.

\section{Conclusion}

In this work, we have presented a prototype-based framework for self-supervised one-shot learning of medical image segmentation tasks. Instead of taking a pre-training representation learning approach, we take a task-learning-based approach. To address the issue of variations in background information between the support and query images, we propose a correlation-based weighting scheme to aggregate the support prototypes according to how related the prototypes are to the query image. Therefore, each pixel on the feature map of the query has a custom prototype. The score for foreground or background is obtained by calculating the cosine similarity of the query feature map pixels with their corresponding prototype. The primary objective of constructing a prototype for each query feature map pixel is to reduce false positives in the predictions by weighing down the contribution of dissimilar prototypes in the final prediction. Despite the limitations of the proposed method, we can see that the proposed method is on par with most contemporary self-supervised segmentation methods.

\bibliographystyle{unsrt}
\bibliography{main}

\begin{thebibliography}{10}
\providecommand{\url}[1]{\texttt{#1}}
\providecommand{\urlprefix}{URL }
\providecommand{\doi}[1]{https://doi.org/#1}

\bibitem{amac2022masksplit}
Amac, M., Sencan, A., Baran, O., Ikizler-Cinbis, N., Cinbis, R.: Masksplit: Self-supervised meta-learning for few-shot semantic segmentation. In: 2022 IEEE/CVF Winter Conference on Applications of Computer Vision (WACV). pp. 428--438. IEEE Computer Society, Los Alamitos, CA, USA (jan 2022). \doi{10.1109/WACV51458.2022.00050}, \url{https://doi.ieeecomputersociety.org/10.1109/WACV51458.2022.00050}

\bibitem{araslanov2021ssaugcon}
Araslanov, N., Roth, S.: Self-supervised augmentation consistency for adapting semantic segmentation. In: 2021 IEEE/CVF Conference on Computer Vision and Pattern Recognition (CVPR). pp. 15379--15389. IEEE Computer Society, Los Alamitos, CA, USA (jun 2021). \doi{10.1109/CVPR46437.2021.01513}, \url{https://doi.ieeecomputersociety.org/10.1109/CVPR46437.2021.01513}

\bibitem{bhunia2019logoret}
Bhunia, A.K., Bhunia, A.K., Ghose, S., Das, A., Roy, P.P., Pal, U.: A deep one-shot network for query-based logo retrieval. Pattern Recognition  \textbf{96},  106965 (2019). \doi{https://doi.org/10.1016/j.patcog.2019.106965}

\bibitem{chen2022apanet}
Chen, J., Gao, B.B., Lu, Z., Xue, J.H., Wang, C., Liao, Q.: Apanet: Adaptive prototypes alignment network for few-shot semantic segmentation. IEEE Transactions on Multimedia pp.~1--1 (2022). \doi{10.1109/TMM.2022.3174405}

\bibitem{chen2019sslmiaicr}
Chen, L., Bentley, P., Mori, K., Misawa, K., Fujiwara, M., Rueckert, D.: Self-supervised learning for medical image analysis using image context restoration. Medical Image Analysis  \textbf{58},  101539 (2019). \doi{https://doi.org/10.1016/j.media.2019.101539}

\bibitem{ding2023crapnet}
Ding, H., Sun, C., Tang, H., Cai, D., Yan, Y.: Few-shot medical image segmentation with cycle-resemblance attention. In: 2023 IEEE/CVF Winter Conference on Applications of Computer Vision (WACV). pp. 2487--2496. IEEE Computer Society, Los Alamitos, CA, USA (jan 2023). \doi{10.1109/WACV56688.2023.00252}, \url{https://doi.ieeecomputersociety.org/10.1109/WACV56688.2023.00252}

\bibitem{Dong2018FewShotSS}
Dong, N., Xing, E.P.: Few-shot semantic segmentation with prototype learning. In: British Machine Vision Conference (2018)

\bibitem{fan2022selfsupp}
Fan, Q., Pei, W., Tai, Y.W., Tang, C.K.: Self-support few-shot semantic segmentation. In: Avidan, S., Brostow, G., Ciss{\'e}, M., Farinella, G.M., Hassner, T. (eds.) Computer Vision -- ECCV 2022. pp. 701--719. Springer Nature Switzerland, Cham (2022)

\bibitem{felzenszwalb2004imgseg}
Felzenszwalb, P.F., Huttenlocher, D.P.: Efficient {Graph-Based} image segmentation. Int. J. Comput. Vis.  \textbf{59}(2),  167--181 (Sep 2004)

\bibitem{gansbeke2021maskcontrast}
Gansbeke, W.V., Vandenhende, S., Georgoulis, S., Gool, L.V.: Unsupervised semantic segmentation by contrasting object mask proposals. In: 2021 IEEE/CVF International Conference on Computer Vision (ICCV). pp. 10032--10042. IEEE Computer Society, Los Alamitos, CA, USA (oct 2021). \doi{10.1109/ICCV48922.2021.00990}, \url{https://doi.ieeecomputersociety.org/10.1109/ICCV48922.2021.00990}

\bibitem{gao2022urltissseg}
Gao, Z., Jia, C., Li, Y., Zhang, X., Hong, B., Wu, J., Gong, T., Wang, C., Meng, D., Zheng, Y., Li, C.: Unsupervised representation learning for tissue segmentation in histopathological images: From global to local contrast. {IEEE} Trans. Medical Imaging  \textbf{41}(12),  3611--3623 (2022). \doi{10.1109/TMI.2022.3191398}, \url{https://doi.org/10.1109/TMI.2022.3191398}

\bibitem{guharoy2020senet}
{Guha Roy}, A., Siddiqui, S., Pölsterl, S., Navab, N., Wachinger, C.: ‘squeeze \& excite’ guided few-shot segmentation of volumetric images. Medical Image Analysis  \textbf{59},  101587 (2020). \doi{https://doi.org/10.1016/j.media.2019.101587}

\bibitem{guizilini2013ossdo}
Guizilini, V., Ramos, F.: Online self-supervised segmentation of dynamic objects. In: 2013 IEEE International Conference on Robotics and Automation. pp. 4720--4727 (2013). \doi{10.1109/ICRA.2013.6631249}

\bibitem{haoyu2021poshp}
He, H., Zhang, J., Thuraisingham, B., Tao, D.: Progressive one-shot human parsing. In: Thirty-Fifth {AAAI} Conference on Artificial Intelligence, {AAAI} 2021, Virtual Event, February 2-9, 2021. pp. 1522--1530. {AAAI} Press (2021), \url{https://ojs.aaai.org/index.php/AAAI/article/view/16243}

\bibitem{hoyer2021threeways}
Hoyer, L., Dai, D., Chen, Y., Koring, A., Saha, S., Gool, L.V.: Three ways to improve semantic segmentation with self-supervised depth estimation. In: 2021 IEEE/CVF Conference on Computer Vision and Pattern Recognition (CVPR). pp. 11125--11135. IEEE Computer Society, Los Alamitos, CA, USA (jun 2021). \doi{10.1109/CVPR46437.2021.01098}, \url{https://doi.ieeecomputersociety.org/10.1109/CVPR46437.2021.01098}

\bibitem{ji2019iic}
Ji, X., Vedaldi, A., Henriques, J.: Invariant information clustering for unsupervised image classification and segmentation. In: 2019 IEEE/CVF International Conference on Computer Vision (ICCV). pp. 9864--9873. IEEE Computer Society, Los Alamitos, CA, USA (nov 2019). \doi{10.1109/ICCV.2019.00996}, \url{https://doi.ieeecomputersociety.org/10.1109/ICCV.2019.00996}

\bibitem{kavur2021chaos}
Kavur, A.E., Gezer, N.S., Barış, M., Aslan, S., Conze, P.H., Groza, V., Pham, D.D., Chatterjee, S., Ernst, P., Özkan, S., Baydar, B., Lachinov, D., Han, S., Pauli, J., Isensee, F., Perkonigg, M., Sathish, R., Rajan, R., Sheet, D., Dovletov, G., Speck, O., Nürnberger, A., Maier-Hein, K.H., {Bozdağı Akar}, G., Ünal, G., Dicle, O., Selver, M.A.: Chaos challenge - combined (ct-mr) healthy abdominal organ segmentation. Medical Image Analysis  \textbf{69},  101950 (2021). \doi{https://doi.org/10.1016/j.media.2020.101950}

\bibitem{li2021asgnet}
Li, G., Jampani, V., Sevilla-Lara, L., Sun, D., Kim, J., Kim, J.: Adaptive prototype learning and allocation for few-shot segmentation. In: 2021 IEEE/CVF Conference on Computer Vision and Pattern Recognition (CVPR). pp. 8330--8339. IEEE Computer Society, Los Alamitos, CA, USA (jun 2021). \doi{10.1109/CVPR46437.2021.00823}, \url{https://doi.ieeecomputersociety.org/10.1109/CVPR46437.2021.00823}

\bibitem{liu2022dpcn}
Liu, J., Bao, Y., Xie, G., Xiong, H., Sonke, J., Gavves, E.: Dynamic prototype convolution network for few-shot semantic segmentation. In: 2022 IEEE/CVF Conference on Computer Vision and Pattern Recognition (CVPR). pp. 11543--11552. IEEE Computer Society, Los Alamitos, CA, USA (jun 2022). \doi{10.1109/CVPR52688.2022.01126}, \url{https://doi.ieeecomputersociety.org/10.1109/CVPR52688.2022.01126}

\bibitem{liu2022fssotmmf}
Liu, W., Zhang, C., Ding, H., Hung, T.Y., Lin, G.: Few-shot segmentation with optimal transport matching and message flow. IEEE Transactions on Multimedia pp. 1--12 (2022). \doi{10.1109/TMM.2022.3187855}

\bibitem{liu2020papnfss}
Liu, Y., Zhang, X., Zhang, S., He, X.: Part-aware prototype network for few-shot semantic segmentation. In: Vedaldi, A., Bischof, H., Brox, T., Frahm, J.M. (eds.) Computer Vision -- ECCV 2020. pp. 142--158. Springer International Publishing, Cham (2020)

\bibitem{okazawa2022ipr}
Okazawa, A.: Interclass prototype relation for few-shot segmentation. In: Avidan, S., Brostow, G., Ciss{\'e}, M., Farinella, G.M., Hassner, T. (eds.) Computer Vision -- ECCV 2022. pp. 362--378. Springer Nature Switzerland, Cham (2022)

\bibitem{ouali2020autouis}
Ouali, Y., Hudelot, C., Tami, M.: Autoregressive unsupervised image segmentation. In: Vedaldi, A., Bischof, H., Brox, T., Frahm, J.M. (eds.) Computer Vision -- ECCV 2020. pp. 142--158. Springer International Publishing, Cham (2020)

\bibitem{ouyang2020superpixels}
Ouyang, C., Biffi, C., Chen, C., Kart, T., Qiu, H., Rueckert, D.: Self-supervision with superpixels: Training few-shot medical image segmentation without annotation. In: Computer Vision -- ECCV 2020. pp. 762--780. Springer International Publishing, Cham (2020)

\bibitem{ouyang2022alpnet}
Ouyang, C., Biffi, C., Chen, C., Kart, T., Qiu, H., Rueckert, D.: Self-supervised learning for few-shot medical image segmentation. IEEE Transactions on Medical Imaging  \textbf{41}(7),  1837--1848 (2022). \doi{10.1109/TMI.2022.3150682}

\bibitem{Rakelly2018ConditionalNF}
Rakelly, K., Shelhamer, E., Darrell, T., Efros, A.A., Levine, S.: Conditional networks for few-shot semantic segmentation. In: International Conference on Learning Representations (2018)

\bibitem{Rakelly2018FewShotSP}
Rakelly, K., Shelhamer, E., Darrell, T., Efros, A.A., Levine, S.: Few-shot segmentation propagation with guided networks. ArXiv  \textbf{abs/1806.07373} (2018)

\bibitem{shaban2017oslss}
Shaban, A., Bansal, S., Liu, Z., Essa, I., Boots, B.: One-shot learning for semantic segmentation. In: British Machine Vision Conference 2017, {BMVC} 2017, London, UK, September 4-7, 2017. {BMVA} Press (2017)

\bibitem{siam2019amp}
Siam, M., Oreshkin, B., Jagersand, M.: Amp: Adaptive masked proxies for few-shot segmentation. In: 2019 IEEE/CVF International Conference on Computer Vision (ICCV). pp. 5248--5257. IEEE Computer Society, Los Alamitos, CA, USA (nov 2019). \doi{10.1109/ICCV.2019.00535}, \url{https://doi.ieeecomputersociety.org/10.1109/ICCV.2019.00535}

\bibitem{MennatullahSiam2019wtimp}
Siam, M., Oreshkin, B.N.: Adaptive masked weight imprinting for few-shot segmentation. In: Workshop at the International Conference on Learning Representations (ICLR) (2019)

\bibitem{singh2018overheadimg}
Singh, S., Batra, A., Pang, G., Torresani, L., Basu, S., Paluri, M., Jawahar, C.V.: Self-supervised feature learning for semantic segmentation of overhead imagery. In: British Machine Vision Conference 2018, {BMVC} 2018, Newcastle, UK, September 3-6, 2018. p.~102. {BMVA} Press (2018), \url{http://bmvc2018.org/contents/papers/0345.pdf}

\bibitem{yang2020dan}
Wang, H., Zhang, X., Hu, Y., Yang, Y., Cao, X., Zhen, X.: Few-shot semantic segmentation with democratic attention networks. In: Vedaldi, A., Bischof, H., Brox, T., Frahm, J.M. (eds.) Computer Vision -- ECCV 2020. pp. 730--746. Springer International Publishing, Cham (2020)

\bibitem{wang2019panet}
Wang, K., Liew, J.H., Zou, Y., Zhou, D., Feng, J.: Panet: Few-shot image semantic segmentation with prototype alignment. In: 2019 IEEE/CVF International Conference on Computer Vision (ICCV). pp. 9196--9205 (2019). \doi{10.1109/ICCV.2019.00929}

\bibitem{wu2022dclass}
Wu, H., Xiao, F., Liang, C.: Dual contrastive learning with anatomical auxiliary supervision for few-shot medical image segmentation. In: Avidan, S., Brostow, G., Ciss{\'e}, M., Farinella, G.M., Hassner, T. (eds.) Computer Vision -- ECCV 2022. pp. 417--434. Springer Nature Switzerland, Cham (2022)

\bibitem{yang2020pmmfss}
Yang, B., Liu, C., Li, B., Jiao, J., Ye, Q.: Prototype mixture models for few-shot semantic segmentation. In: Vedaldi, A., Bischof, H., Brox, T., Frahm, J.M. (eds.) Computer Vision -- ECCV 2020. pp. 763--778. Springer International Publishing, Cham (2020)

\bibitem{zhang2021sgcgfss}
Zhang, B., Xiao, J., Qin, T.: Self-guided and cross-guided learning for few-shot segmentation. In: 2021 IEEE/CVF Conference on Computer Vision and Pattern Recognition (CVPR). pp. 8308--8317. IEEE Computer Society, Los Alamitos, CA, USA (jun 2021). \doi{10.1109/CVPR46437.2021.00821}, \url{https://doi.ieeecomputersociety.org/10.1109/CVPR46437.2021.00821}

\bibitem{zhang2020guidednetcrf}
Zhang, K., Zheng, Y., Deng, X., Jia, W., Lian, J., Chen, X.: Guided networks for few-shot image segmentation and fully connected crfs. Electronics  \textbf{9}(9) (2020). \doi{10.3390/electronics9091508}, \url{https://www.mdpi.com/2079-9292/9/9/1508}

\bibitem{zhang2020sgone}
Zhang, X., Wei, Y., Yang, Y., Huang, T.S.: Sg-one: Similarity guidance network for one-shot semantic segmentation. IEEE Transactions on Cybernetics  \textbf{50}(9),  3855--3865 (2020). \doi{10.1109/TCYB.2020.2992433}

\bibitem{Zhu2020SelfSupervisedTF}
Zhu, K., Zhai, W., Zha, Z., Cao, Y.: Self-supervised tuning for few-shot segmentation. In: International Joint Conference on Artificial Intelligence (2020), \url{https://api.semanticscholar.org/CorpusID:215744862}

\end{thebibliography}



\end{document}